\documentclass[letterpaper, 10 pt, conference]{ieeeconf}  % Comment this line out if you need a4paper

\IEEEoverridecommandlockouts                              % This command is only needed if 
\usepackage{graphics} % for pdf, bitmapped graphics files
\usepackage{graphicx}

\usepackage{epsfig} % for postscript graphics files
\usepackage{amsmath} % assumes amsmath package installed
\usepackage{amssymb}  % assumes amsmath package installed
 \usepackage{algorithm}
\usepackage[compatible]{algpseudocode}
\algnewcommand\AAND{\textbf{ and }}
\algnewcommand\Or{\textbf{ or }}
\usepackage{color}
\usepackage{citesort}
\usepackage{flushend}
\usepackage{url}
\usepackage[table,xcdraw]{xcolor}

\usepackage[colorinlistoftodos]{todonotes}

\DeclareMathAlphabet{\pazocal}{OMS}{zplm}{m}{n}

\newcommand{\Vs}{\pazocal{V}}

\newcommand{\Is}{\pazocal{I}}

\DeclareMathAlphabet{\mathpzc}{OT1}{pzc}{m}{it}

\usepackage{array}
\newcolumntype{C}[1]{>{\centering\arraybackslash}p{#1}}
\newcolumntype{M}[1]{>{\raggedright\arraybackslash}p{#1}}

\usepackage{array} 
\newcolumntype{L}[1]{>{\raggedright\let\newline\\\arraybackslash\hspace{0pt}}m{#1}}	
\newcolumntype{S}[1]{>{\centering\let\newline\\\arraybackslash\hspace{0pt}}m{#1}}
\newcolumntype{R}[1]{>{\raggedleft\let\newline\\\arraybackslash\hspace{0pt}}m{#1}}

\def \rmfname{RMF-Owl}
\makeatletter
\renewcommand*{\@opargbegintheorem}[3]{\trivlist
  \item[\hskip \labelsep{\itshape #1\ #2}] \textit{(#3)}\ }
\makeatother
\title{\LARGE \bf
RMF-Owl: A Collision-Tolerant Flying Robot for Autonomous Subterranean Exploration
}

\author{Paolo De Petris$^1$, Huan Nguyen$^1$, Mihir Dharmadhikari$^2$, Mihir Kulkarni$^2$,\\ Nikhil Khedekar$^1$, Frank Mascarich$^2$, and Kostas Alexis$^1$% <-this % stops a space
\thanks{
This material is based upon work supported by a) the Defense Advanced Research Projects Agency (DARPA) under Agreement No. HR00111820045, and b) the Research Council of Norway project SENTIENT, grant number 321435. The presented content and ideas are solely those of the authors.
}
\thanks{$^1$The authors are with the Autonomous Robots Lab, NTNU, O.S. Bragstads Plass 2D , 7034, Trondheim, NO
$^2$The authors are with the University of Nevada, Reno, 1664 N. Virginia, 89557, Reno, NV, USA
        {\tt\small paolo.de.petris@ntnu.no}}%
}

\begin{document}

\maketitle
\thispagestyle{empty}
\pagestyle{empty}

%%%%%%%%%%%%%%%%%%%%%%%%%%%%%%%%%%%%%%%%%%%%%%%%%%%%%%%%%%%%%%%%%%%%%%%%%%%%%%%%
\begin{abstract}
This work presents the design, hardware realization, autonomous exploration and object detection capabilities of \rmfname, a new collision-tolerant aerial robot tailored for resilient autonomous subterranean exploration. The system is custom built for underground exploration with focus on collision tolerance, resilient autonomy with robust localization and mapping, alongside high-performance exploration path planning in confined, obstacle-filled and topologically complex underground environments. Moreover, \rmfname~ offers the ability to search, detect and locate objects of interest which can be particularly useful in search and rescue missions. A series of results from field experiments are presented in order to demonstrate the system's ability to autonomously explore challenging unknown underground environments.

\end{abstract}

%%%%%%%%%%%%%%%%%%%%%%%%%%%%%%%%%%%%%%%%%%%%%%%%%%%%%%%%%%%%%%%%%%%%%%%%%%%%%%%%
\section{INTRODUCTION}\label{sec:intro}
\begin{figure}[h!]
\centering
    \includegraphics[width=0.88\columnwidth]{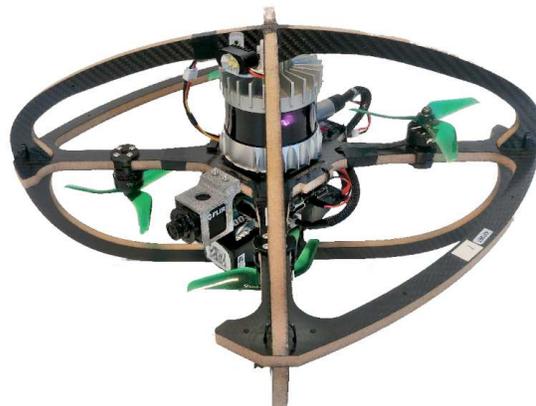}
\vspace{-3ex} 
\caption{\rmfname: The collision-tolerant flying system designed for resilient autonomous subterranean exploration.}
\label{fig:rmfintrophoto}
\vspace{-3ex} 
\end{figure}

Research in aerial robotics has presented exciting progress towards fast and agile navigation combined with accurate localization, precise mapping, and collision avoidance. This has allowed the execution of autonomous complex missions such as exploration of unknown subterranean environments~\cite{GBPLANNER_JFR_2020,NIR_ICUAS_2017}, infrastructure inspection~\cite{sa2014vertical,bircher2017incremental}, search and rescue~\cite{tomic2012toward} and more. However, finding the maneuver to avoid all possible obstacles in the environment, or identifying the way to fit and fly through extremely confined settings is particularly challenging and, sometimes, not always possible. The problem is further amplified when one considers the implicit uncertainty of the robot localization and mapping especially in GNSS-denied sensor-degraded environments, as well as control inaccuracies. The more we push the limits of the flight envelope and the scope of autonomous missions, the harder the problem becomes for the onboard localization, mapping and collision-free navigation processes. The above motivated the process for the design and autonomy functionalities of the presented \rmfname~aerial robot, shown in Figure~\ref{fig:rmfintrophoto}, which is tailored to subterranean exploration.

%%%%%%%%%%%%%%%%%%%%%%%%%%%%%%
% \subsection{Challenges in Subterranean Environments}
%%%%%%%%%%%%%%%%%%%%%%%%%%%%%%
Some of the greatest hazards on Earth can be found in subterranean environments: tunnel collapses, limited access to food and oxygen, maze-like cave systems, and oppressing darkness all represent serious dangers to those who work or operate underground. Utilization of autonomous robotic technologies to undertake significant workload is thus essential. Focusing specifically on the task of autonomous exploration and mapping of such settings and motivated by the DARPA Subterranean Challenge~\cite{darpasubt} activities, the operational domains can be distinguished in a) Tunnel-like environments such as underground mines, b) Urban underground infrastructure such as the subway network in metropolitan cities, and c) Cave networks often involving highly irregular geological features and unpredictable topologies. Aiming to support the goal of autonomous exploration in diverse subterranean settings, we develop and present \rmfname, an aerial robot tailored to navigate safely within narrow geometries, quickly within vast spaces, and seamlessly across complex topologies. The system is capable of resilient onboard localization and mapping, as well as autonomous exploration path planning. To optimize for its highest prioritized task, that of exploring through extremely narrow and obstacle-filled subterranean passages, \rmfname~employs a collision-tolerant design to maximize its resilience. In this work, we detail the airframe and mechatronic design, outline the onboard localization and mapping system as well as the pipeline to detect and localize objects of interest, summarize the onboard exploration planner and present selected experimental results from underground environments.

The remainder of this paper is structured as follows: Section~\ref{sec:related} outlines relevant work in the domain of collision-tolerant flying robots. The new robot's design is presented in Section~\ref{sec:sys_des}, followed by an overview of the realized autonomy stack in Section~\ref{sec:res_aut_nav}. Field experimental results are detailed in~\ref{sec:fie_exp}. Finally, conclusions are drawn in Section~\ref{sec:concl}.

%%%%%%%%%%%%%%%%%%%%%%%%%%%%%%%%%%%%%%%%%%%%%%%%%%%%%%%%%%%%%%%%%%%%%%%%%%%%%%%%
\section{RELATED WORK}\label{sec:related}
A niche community of researchers has worked in the domain of collision-tolerant flying robots. An overview of the literature allows for the derivation of a taxonomy of designs with respect to the way collision-tolerance is implemented and how it is exploited. Considering rigid frame designs, a set of researchers utilize rolling cage frames to support safety post-collision~\cite{briod2014collision,salaan2019development,atay2021spherical}. Focusing on enabling a dual modality of navigation, the authors in~\cite{yamada2017development} propose a caged robot with the cage designed to double as a wheel on the ground. Optimizing for reduced weight and mechanical complexity, rigid designs not involving a rolling cage are presented in~\cite{RMF_ICRA2021,sabet2019rollocopter,moon2018uni} including previous work from the authors. Considering the potential benefits of a compliant design in best absorbing the impact effects, a set of relevant designs have been proposed and follow both biomimetic and more classical engineered concepts~\cite{liu2021toward,mintchev2017insect,klaptocz2013euler}. Focusing on particularly lightweight protective designs, the authors in~\cite{sareh2018rotorigami,kornatowski2017origami,sareh2018spinning} present novel origami shroud designs. Emphasizing both compliance and the ability to re-orient in case of a crash and fall on the ground, the works in~\cite{zha2020collision,de2021being} present tensegrity-based soft collision-tolerant flying robots. Departing from multirotor and broadly rotorcraft designs, the contributions in~\cite{klaptocz2010indoor,vourtsis2021robotic} present collision-tolerant fixed-wing designs including bioinspired shape-morphing for higher protection when not flying. Considering flapping-wing designs and miniaturized systems, a set of relevant collision-tolerant systems are presented in~\cite{phan2020mechanisms,chen2021collision}. Beyond contributions in collision-tolerant aerial robot designs, a set of works focus on specialized autonomy functionalities exploiting impact resilience. The efforts in~\cite{zha2021exploiting,mote2016framework} propose path planning methods that are aware and exploit the resilience introduced by collision-tolerant frames. Considering the potential of using collisions as means of sensing, a set of relevant contributions are presented in~\cite{liu2021sensing,lew2019contact,mulgaonkar2020tiercel} targeting to acquire both odometry and map information. Emphasizing multi-robot autonomy, the authors in~\cite{mulgaonkar2017robust} enable robust swarming without explicit collision avoidance mechanisms. Within the spectrum of the presented taxonomy of collision resilient aerial robots, this work presents a rigid non-rotating design with a particularly lightweight design that focuses on advanced perception and autonomy features tailored to subterranean exploration.

%%%%%%%%%%%%%%%%%%%%%%%%%%%%%%%%%%%%%%%%%%%%%%%%%%%%%%%%%%%%%%%%%%%%%%%%%%%%%%%%
\section{SYSTEM DESIGN}\label{sec:sys_des}
This section overviews the design concept of \rmfname. \rmfname~is part of the Resilient Micro Flyer (RMF) aerial robot family with a first more lightweight system presented in~\cite{RMF_ICRA2021} and a version tailored to nuclear radiation characterization detailed in~\cite{RADMF_ICRA_2021}.

%%%%%%%%%%%%%%%%%%%%%%%%%%%%%%%%%%%%%%%%%%%%%%%%%%%
\subsection{Airframe}
%%%%%%%%%%%%%%%%%%%%%%%%%%%%%%%%%%%%%%%%%%%%%%%%%%%
The frame of \rmfname~is designed for prolonged endurance, agile flight, light weight and collision-tolerance.
The main rigid component is fabricated using carbon-foam sandwich material (total width: $10\textrm{mm}$, with $0.75\textrm{mm}$ carbon on each side. The carbon and foam core densities are approximately $0.0011 \textrm{g}/\textrm{mm}^3$ and $0.00008758\textrm{g}/\textrm{mm}^3$ respectively, leading to a total airframe weight of $145\textrm{g}$. 
The decision of leaving the front and the back open is to avoid obstruction with respect to the field of view of the front camera and respect symmetry.
The platform integrates four T-Motor F60PRO IV V2.0 KV1950 DC Brushless motors controlled through their electronic speed controllers.
Moreover, \rmfname~ integrates a PixRacer R15 as its main low-level autopilot unit offering attitude and thrust control. High-level position control and navigation autonomy is facilitated through a different processing board as detailed further below. The resulting dimensions of \rmfname, are  $38\times38\times24 \textrm{cm}~(\textit{L}\times\textit{W}\times\textit{H})$ and the total weight, including all the sensing and processing components, cabling and its battery is $1.45\textrm{kg}$. In this configuration the power to weight ratio is equal to 2 and the resulting flight time is 10 minutes.

%%%%%%%%%%%%%%%%%%%%%%%%%%%%%%%%%%%%%%%%%%%%%%%%%%%
\subsection{Electronics}
\begin{figure}[h!]
\vspace{-2ex} 
\centering
     \includegraphics[width=0.95\columnwidth]{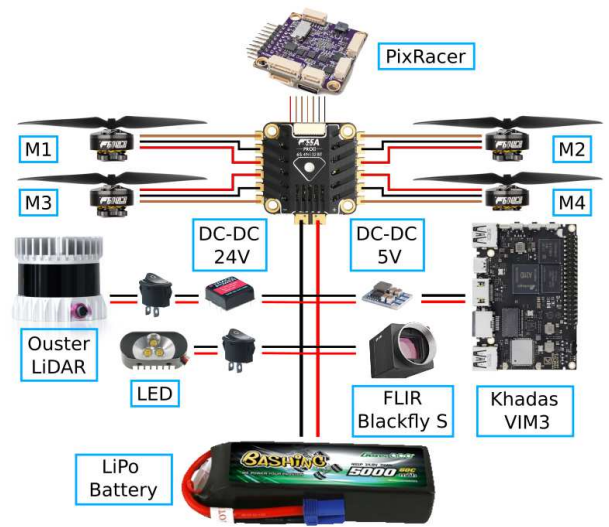}
\vspace{-2ex} 
\caption{Overview of the \rmfname\ electronics.}
\label{fig:electronics}
\vspace{-2ex} 
\end{figure}

%%%%%%%%%%%%%%%%%%%%%%%%%%%%%%%%%%%%%%%%%%%%%%%%%%%
As depicted in Figure~\ref{fig:electronics}, \rmfname\ is powered via a single Spektrum 4s 5000 mAh LiPo battery. When fully charged, this provides 16.8 V directly to the Power Distribution Board (PDB), a T-Motor F55A Pro II 4-in-1 ESC. Two separate lines are derived from the battery circuitry. The first powers an Ouster OS0 LiDAR sensor, with the usage of a Tracopower 30W Isolated DC-DC Converter (24V). The second, with the addition of a 5V/3A DC-DC, provides power to the onboard computer, described in detail below. The PDB delivers the proper power to all of the motors of \rmfname\ and to the autopilot, via a separate 5V built-in line. To enable visual detection in low-light/dark underground environments, the battery also powers a Lumenier sUAS High Lumen Overt LED, triggered via a PWM line directly from the autopilot to reduce power consumption and thus increase efficiency.

%%%%%%%%%%%%%%%%%%%%%%%%%%%%%%%%%%%%%%%%%%%%%%%%%%%
\subsection{Frame analysis under collision}
%%%%%%%%%%%%%%%%%%%%%%%%%%%%%%%%%%%%%%%%%%%%%%%%%%%
In order to evaluate the mechanical properties of the designed carbon-foam frame, two studies are conducted. Firstly, a CAD simulation is conducted in order to identify possible breaking points on the structure and offer input for possible future designs. Guided jointly by intuition and by the results of this first analysis, we then utilized duplicated parts of the current design, as well as a previous version of the frame - albeit thinner and thus a bit weaker -  to get an understanding of what is the order of magnitude of forces that the frame can sustain and verify where the breaking points are to be expected. 

\subsubsection{CAD Simulation}
\begin{figure}[h!]
\centering
    \includegraphics[width=0.99\columnwidth]{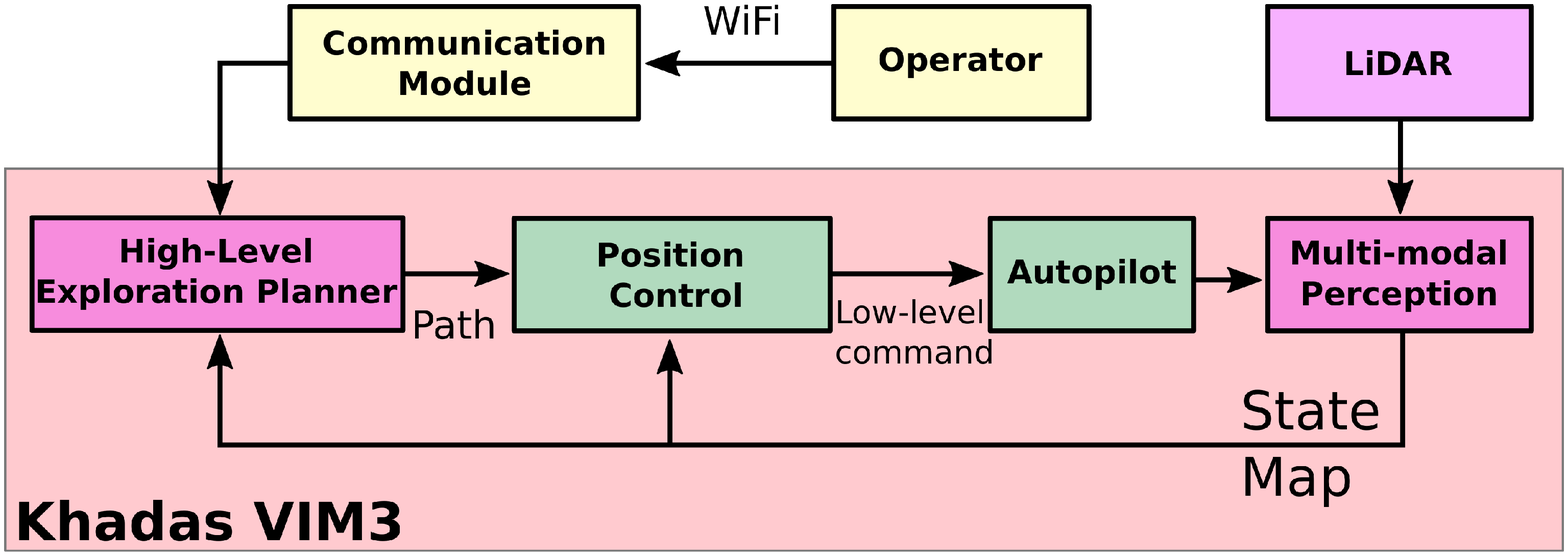}
\caption{Schematic overview of the core functional modules of \rmfname. The components inside the red shaded area are executed onboard the robot. Green color depicts robot-specific functionalities} 
% this caption is taken from the phaseII report, must be adapted to be coherent with the text flow
% , while magenta is used for software components that are unified across the robots of Team CERBERUS~\cite{cerberuswebsite}. As such \rmfname\ is integrated to a large apparatus of systems and methods tailored to subterranean autonomy.

\label{fig:architecture}
\vspace{-2ex} 
\end{figure}

In this test we applied different static forces to the model in simulation in an iterative fashion, looking for breaking points and possible regions that can be improved in next design iterations. 
As can be seen in Figure~\ref{fig:architecture}, the selected applied pressure points are top, bottom, lateral, frontal straight and frontal $45\textit{°}$. The Figure allows to derive an understanding as to which locations of the frame experience the maximum stress for certain directions of collision forces. The selection of collision points reflects the types of impact most commonly encountered during flight. 
% Table~\ref{tab:cadtest} reports selected significant values of overall frame displacement output from the simulation given different values of forces applied.

\begin{figure}[h!]
\centering
    \includegraphics[width=0.94\columnwidth]{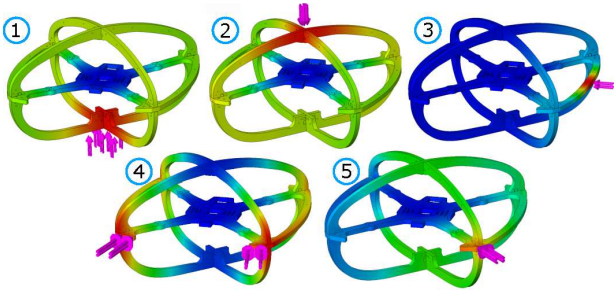}
\vspace{-2ex}
\caption{CAD stress test-based qualitative analysis conducted on the frame of \rmfname\.\ The static forces, represented by the pink arrows, are iteratively applied on the bottom $(1)$, top $(2)$, side $(3)$, front $(4)$ and front with an angle of $45\textit{°}$ $(5)$. The resulting frame displacement (reflecting stress) ranges from a minimum (blue) to a maximum value (red). Even in cases where such CAD simulations can present numerical errors in their results, the locations where maximum stress is applied and thus breaking is to be anticipated are expected to be qualitatively correct.}
\label{fig:cad_test}
\vspace{-2ex} 
\end{figure}

% \begin{table}[h!]
% \begin{tabular}{cccccc}
% \multicolumn{1}{l}{}                           & \multicolumn{5}{c}{\textbf{Overall Displacement {[}mm{]}}}               \\
% \rowcolor[HTML]{F3F3F3} 
% \cellcolor[HTML]{EFEFEF}\textbf{Force {[}N{]}} & \textbf{(1)} & \textbf{(2)} & \textbf{(3)} & \textbf{(4)} & \textbf{(5)} \\
% \textbf{5}                                     & 4.3344       & 4.3893       & 1.479        & 0.884762     & 0.038506     \\
% \textbf{10}                                    & 10.8359      & 10.9732      & 3.698        & 2.2119       & 0.096266     \\
% \textbf{50}                                    & 54.1795      & 54.8661      & 18.49        & 11.0595      & 0.48133      \\
% \textbf{100}                                   & 108.36       & 109.73       & 37.96        & 11.06        & 0.96        
% \end{tabular}
% \caption{CAD Simulation results of frame displacement.}\label{tab:cadtest}
% \end{table}

\subsubsection{Hydraulic Press Test}
\begin{figure}[h!]
\centering
    \includegraphics[width=0.99\columnwidth]{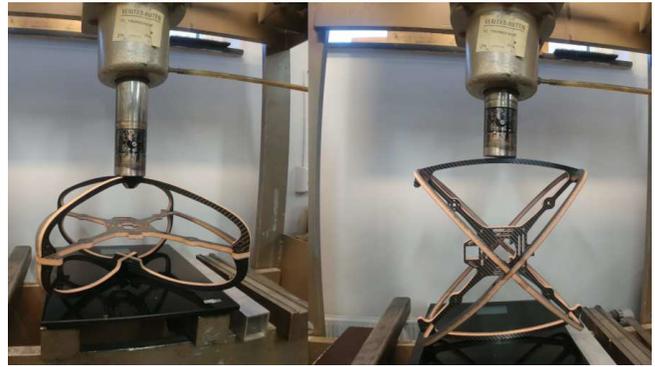}
\vspace{-3ex}
\caption{Hydraulic press test experimental setup for case (iii) and (iv).}
\label{fig:presstest}
\vspace{-2ex} 
\end{figure}

In this second test the goal was to find the maximum force that the frame can sustain before breaking. Accordingly, a set of tests were conducted by placing: a duplicate part of the bottom of the frame $(i)$, a duplicate part of the frame lateral protection $(ii)$, and a previous - thinner - version of the frame in a normal operating position $(iii)$ and also rolled it by $90\textit{°}$ $(iv)$ in between a piston of an industrial hydraulic press and a scale. An indicative setup example is depicted in Figure~\ref{fig:presstest}. Even though inaccuracies are introduced by the use of a frame with thinner carbon-foam arms and surrounding shroud and the utilized scale to estimate the force can present noise, this test allows to a) identify the approximate magnitude of forces that the frame can sustain, and b) verify the locations of maximum stress as also derived by the simulation studies. The maximum forces recorded during each of these experiments - as noted right before the frame broke - are reported in Table~\ref{tab:presstest}. %As the \rmfname~utilizes a thicker carbon-foam design it can sustain at least the presented forces during impact. 

\begin{table}[h!]
\centering

\begin{tabular}{llll}
\rowcolor[HTML]{EFEFEF} 
\multicolumn{1}{c}{\cellcolor[HTML]{EFEFEF}\textbf{(i)}} & \multicolumn{1}{c}{\cellcolor[HTML]{EFEFEF}\textbf{(ii)}} & \multicolumn{1}{c}{\cellcolor[HTML]{EFEFEF}\textbf{(iii)}} & \multicolumn{1}{c}{\cellcolor[HTML]{EFEFEF}\textbf{(iv)}} \\
\multicolumn{1}{c}{285N}                                 & \multicolumn{1}{c}{520N}                                  & \multicolumn{1}{c}{285N}                                   & \multicolumn{1}{c}{128N}                                  
\end{tabular}
%\vspace{-1ex}
\caption{Maximum force recorded before breaking in each of the experiments with the hydraulic press.}\label{tab:presstest}
%\vspace{-1ex}
\end{table}

%%%%%%%%%%%%%%%%%%%%%%%%%%%%%%%%%%%%%%%%%%%%%%%%%%%
\subsection{High-Level Sensing and Processing Payload} \label{high_level}
%%%%%%%%%%%%%%%%%%%%%%%%%%%%%%%%%%%%%%%%%%%%%%%%%%%
The sensing payload of \rmfname\ is tailored to the goal of resilient autonomy in subterranean environments, while maintaining a very lightweight configuration.
The system integrates a sensing suite consisting of an Ouster OS0-64 LiDAR sensor with Field Of View (FOV) $FOV=[360,90]^\circ$ and a Flir Blackfly S color camera with $FOV=[85,64]^\circ$, interfaced with a Khadas VIM3 Pro Single Board Computer (SBC) incorporating $\times 4$ 2.2Ghz Cortex-A73 cores, paired with $\times2$ 1.8Ghz Cortex-A53 cores implementing an A311D big-little architecture, alongside a Neural Processing Unit (NPU) offering $5.0$ TOPS for dedicated neural network inference. The Khadas SBC interfaces the PixRacer autopilot to which it provides navigation commands. The SBC is also responsible for running all the localization and mapping, $3\textrm{D}$ occupancy mapping, path planning, as well as object detection and localization algorithms towards autonomous exploration in GPS-denied and confined environments.

%%%%%%%%%%%%%%%%%%%%%%%%%%%%%%%%%%%%%%%%%%%%%%%%%%%%%%%%%%%%%%%%%%%%%%%%%%%%%%%%
\section{RESILIENT AUTONOMOUS EXPLORATION}\label{sec:res_aut_nav}
\rmfname\ implements a comprehensive autonomy stack that facilitates autonomous exploration and object search capabilities in complex subterranean environments. 

%%%%%%%%%%%%%%%%%%%%%%%%%%%%%%%%%%%%%%%%%%%%%%%%%%%
\subsection{Architecture}
%%%%%%%%%%%%%%%%%%%%%%%%%%%%%%%%%%%%%%%%%%%%%%%%%%%
\rmfname~uses a high-level exploration path planner, described in Section~\ref{ss:planning}, to plan feasible paths for the robot. The path is then tracked by a Proportional Integral Derivative (PID) controller described in Section~\ref{ss:control}  and the low-level attitude and thrust commands are executed by the onboard autopilot. The odometry feedback of the robot and the map of its environment are provided by the localization and mapping solution presented in Section~\ref{ss:perception}.  

% \todo{Figure requires small modification: a) add block ``LiDAR'' going into the slam block, b) we can just write Position Control as it involves the carrot too.}

%%%%%%%%%%%%%%%%%%%%%%%%%%%%%%%%%%%%%%%%%%%%%%%%%%%
\subsection{Perception} \label{ss:perception}
%%%%%%%%%%%%%%%%%%%%%%%%%%%%%%%%%%%%%%%%%%%%%%%%%%%
The operational profile of \rmfname~requires robust localization and mapping in subterranean environments. To this end we utilize the LiDAR odometry and mapping component of CompSLAM~\cite{CompSLAM} which relies on the core ideas in~\cite{loam}. The method utilizes $10\textrm{Hz}$ point cloud readings from the onboard OS0-64 LiDAR. At full rate, it performs a scan-to-scan matching step, while at half rate ($5\textrm{Hz}$) it performs scan-to-submap matching. Accordingly, the robot estimates its pose $\hat{\mathbf{p}}_k$ in the environment and simultaneously reconstructs a pointcloud model of the map $\hat{\mathbf{m}}_k$ of the environment. The map is maintained in a $1010\times1010\times510\textrm{m}$ cuboid around the initial position of the robot. If the robot gets too close to a face of the cuboid map, the entire map is shifted in a direction normal to the face such that the robot is always well contained within the map. To deal with computational limitations, the submap used for the scan-to-submap step has a maximum size of $110\times 110\times 110\textrm{m}$. For the scan-to-scan alignment step, $768$ sharp corners and $1536$ flat surface features are considered for each pointcloud observation. To handle very large environments, the software has the option to utilize only $32$ of the $64$ channels of the OS0-64 sensor thus reducing the computational cost. The reliable $5\textrm{Hz}$ pose update $\hat{\mathbf{p}}_k$ is then fused with the onboard IMU using an Extended Kalman Filter based on the Multi-Sensor Fusion (MSF) framework~\cite{lynen13robust}. The latter allows to acquire a smooth high frequency ($100\textrm{Hz}$) estimate $\hat{\mathbf{p}}_\ell^\prime$ which is then employed by the robot's position controller described in Section~\ref{ss:control}. 

Beyond the requirement for robust localization and mapping, \rmfname~also builds a volumetric map that is used to facilitate autonomous path planning, as well as object localization. By associating the incoming point cloud observations $\textbf{o}_k$ with the related pose estimates $\hat{\mathbf{p}}_k$, the system incrementally builds an occupancy map $\hat{\boldsymbol{\mu}}_k$ using Voxblox~\cite{voxblox}, a computationally efficient incremental mapping framework. For most missions, a voxel edge size of $0.2\textrm{m}$ is utilized.

\subsection{Exploration Path Planning} \label{ss:planning}
% \todo{Ask MihirD to write about this}

% \textcolor{red}{What we need here:}
% \begin{itemize}
%     \item Executive summary of GBPlanner functionalities - both local and global (best done via a diagram)
%     \item Comment on how we obviously can benefit from collision-tolerance in terms of modeling the robot size more aggressively
    
% \end{itemize}

\begin{figure}[h!]
\centering
    \includegraphics[width=0.99\columnwidth]{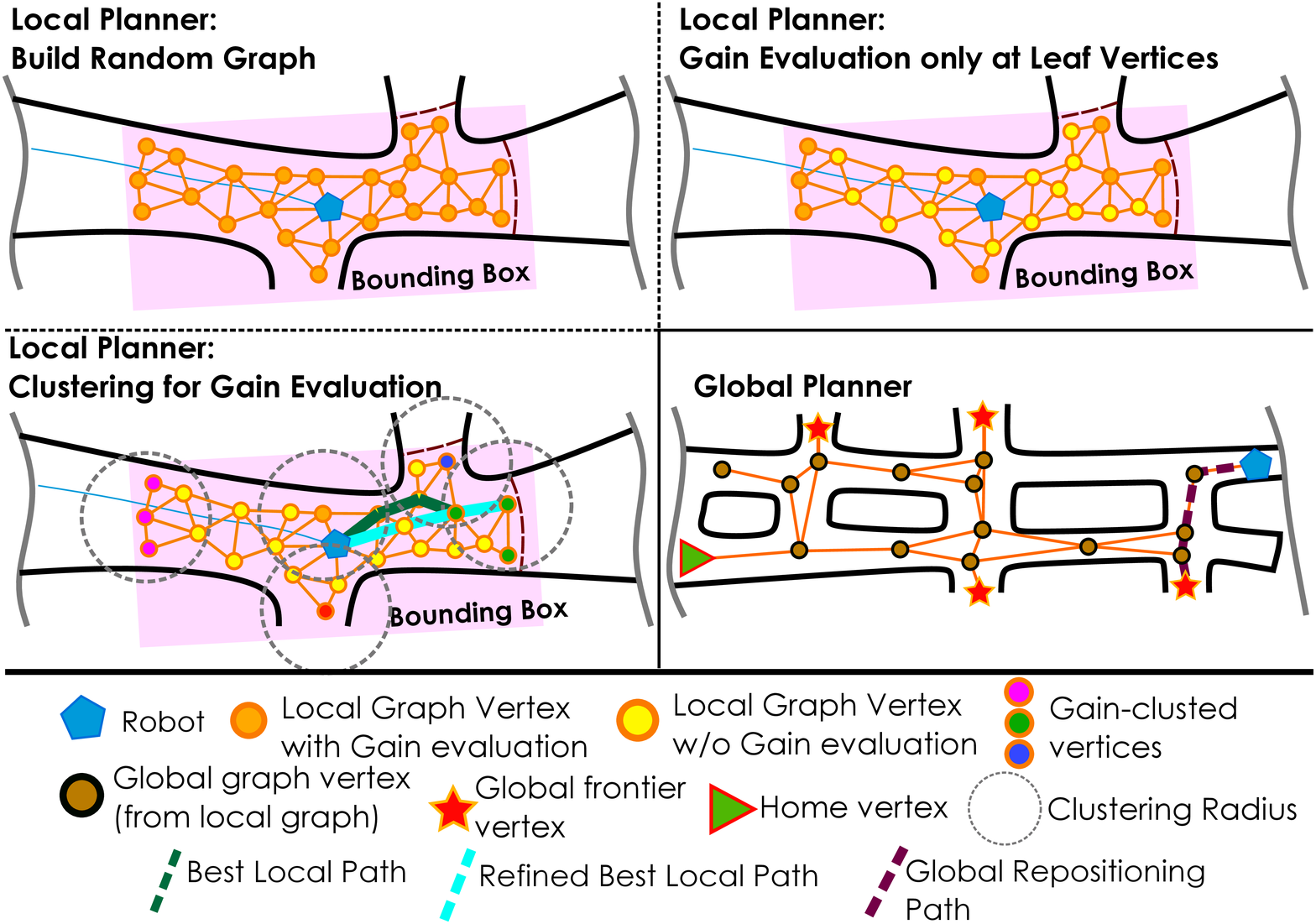}
\caption{Overview of GBPlanner2 functionalities.} 
% this caption is taken from the phaseII report, must be adapted to be coherent with the text flow

\label{fig:gbplanner_overview}
\vspace{-2ex} 
\end{figure}

To achieve autonomous exploration, \rmfname~utilizes a graph-based path planner called GBPlanner2~\cite{GBPlanner2_ArxivVersion}. The method builds upon our previous work~\cite{GBPLANNER_JFR_2020} and uses a bifurcated local/global planning architecture. The local planner is responsible for identifying efficient exploration paths that respect the robot motion and perception constraints. On the other hand, the global planner provides functionality for a) re-positioning the robot towards a previously perceived frontier of the exploration space when the local planner reports inability to find a path of significant exploration gain, and b) ensuring that the robot returns home within its endurance limits. Figure~\ref{fig:gbplanner_overview} provides an overview of the functionalities of GBPlanner2.

In further detail, the local exploration planner of GBPlanner2 utilizes the Voxblox map $\hat{\boldsymbol{\mu}}_k$ to first sample vertices in free space within a local bounding box that is adaptively calculated using the aggregated point cloud around the current robot location. These vertices are connected by collision-free edges to build an undirected graph. Subsequently, the Dijkstra's algorithm is used to calculate the shortest paths starting from the current robot location. For each vertex $\nu_i$ in the graph, an information gain, called $\textbf{VolumetricGain}(\nu_i) =\epsilon_G \hat{\boldsymbol{\mu}}_{k,unk}^{\nu_i}$ is calculated, where $\hat{\boldsymbol{\mu}}_{k,unk}^{\nu_i} \subset \hat{\boldsymbol{\mu}}_k$ is the unknown part of the occupancy map that would lie in the robot's modeled sensor frustum if the robot was at the vertex $\nu_i$ and $\epsilon_G>0$ is a tunable weight. This gain is then accumulated over a path to derive the total gain, called \textbf{ExplorationGain}, of that path. Since this is among the most computationally demanding processes - as it involves extensive raycasting - the planner provides the functionality to calculate the gain only at the leaf vertices of the shortest paths. Furthermore, the leaf vertices can be clustered using a radius $\rho>0$, allowing to approximate the gain of some vertices based on the calculated gain of a nearby vertex. These approximations are necessary for computationally constrained Micro Aerial Vehicles like \rmfname~to operate in large environments. Finally, the path having the highest \textbf{ExplorationGain} is selected, improved for safety as described in~\cite{GBPLANNER_JFR_2020}, and commanded as a reference to the position controller.

% An information gain, called \textbf{VolumetricGain}, that relates to the expected unknown volume to be mapped by the robot's sensor(s) from a vertex, is calculated for the vertices in the graph and accumulated over a path to derive the total gain, called \textbf{ExplorationGain}, of that path. 
%  and $\epsilon_G$ is a tunable weight
% The total accumulated gain over a path, called \textbf{ExplorationGain}, of that path

% It is noted that GBPlanner2 employs the Voxblox map $\hat{\boldsymbol{\mu}}_k$ as its volumetric mapping representation for collision checking and volumetric calculation. 
GBPlanner2 models the robot as a cuboid. A vertex is said to lie in free space if considering the robot's cuboid on it then all its voxels are free. However, this leads to sub-optimal behavior in narrow environments due to the map resolution. When the cuboid size is not a multiple of the map resolution along any axis, the actual volume checked for collision is larger than the size of the cuboid. Due to computational constraints, the map resolution used for \rmfname~is relatively low ($0.2-0.25\textrm{m}$ in most missions) and comparable to the size of the robot, as detailed in Section~\ref{sec:sys_des}. Hence, the above issue is prominent. To reliably traverse very narrow passages we exploit the system's collision tolerance to optimistically set the size of the modelled cuboid without any additional safety margins.

The global layer of GBPlanner2 maintains a sparse global graph built by incrementally appending only the high exploration gain paths from each local planning step, as well as the robot's state. The vertices in this graph that have a volumetric gain higher than a set threshold are characterized as ``frontier vertices''. When the local layer is unable to find a path with an \textbf{ExplorationGain} higher than a threshold for $k$ consecutive iterations, the global planner is triggered to re-position the robot to one of the frontier vertices. The frontier vertex to re-position to is selected based on the frontier's volumetric gain and the exploration time remaining after reaching that frontier~\cite{GBPLANNER_JFR_2020}. Furthermore, in each local planning iteration, the global planner finds the shortest path to the home location and commands it to the robot if the remaining endurance is only sufficient for it to return home.

%%%%%%%%%%%%%%%%%%%%%%%%%%%%%%%%%%%%%%%%%%%%%%%%%%%
\subsection{Control} \label{ss:control}
%%%%%%%%%%%%%%%%%%%%%%%%%%%%%%%%%%%%%%%%%%%%%%%%%%%
The position controller of \rmfname~is a fixed-gain PID scheme, while the yaw controller utilizes a proportional scheme. This controller interfaces the low-level autopilot of the system which provides roll and pitch reference tracking, yaw rate reference tracking and thrust control. Let $\Is$ be the inertial frame, and $\Vs$ the yaw-rotated inertial frame. The outputs of the position and yaw controllers are the commanded acceleration vectors expressed in $\Is$, $[{}^{\Is}a^x_r, {}^{\Is}a^y_r, {}^{\Is}a^z_r]$, and yaw rate $\dot{\psi}_r$ which are then converted to the attitude-thrust command, as per \cite{michael2011cooperative}, and forwarded to the low-level controller inside the autopilot running the PX4 stack:

\scriptsize
\begin{eqnarray} \label{eq:pos_yaw_control}
{}^{\Is}a^x_r &=& K^{x}_P(x_r-x) + K^{x}_I\left.\begin{matrix}
\int (x_r-x)dt
\end{matrix}\right|_{I_{\min}^{x}}^{I_{\max}^{x}} + K^{x}_D (\dot{x}_r - \dot{x}) \\ \nonumber  
{}^{\Is}a^y_r &=& K^{y}_P(y_r-y) + K^{y}_I\left.\begin{matrix}
\int (y_r-y)dt
\end{matrix}\right|_{I_{\min}^{y}}^{I_{\max}^{y}} + K^{y}_D (\dot{y}_r - \dot{y}) \\ \nonumber
{}^{\Is}a^z_r &=& K^{z}_P(z_r-z) + K^{z}_I\left.\begin{matrix}
\int (z_r-z)dt
\end{matrix}\right|_{I_{\min}^z}^{I_{\max}^z} + K^{z}_D (\dot{z}_r - \dot{z}) \\ \nonumber
\dot{\psi}_r &=& K^{\psi}_P (\psi_r - \psi)
\end{eqnarray}
\normalsize
where $[x_r, y_r, z_r, \psi_r], [x, y, z, \psi]$ are the reference and estimated position and yaw angle of the robot, respectively expressed in $\Is$, $I_{\min}^{j},I_{\max}^j,~j\rightarrow x,y,z$ are the saturation minimum and maximum values of the control integrals, and $K^{j}_P,K^{j}_I,K^{j}_D,~j\rightarrow x,y,z,\psi$ are the control gains.

%%%%%%%%%%%%%%%%%%%%%%%%%%%%%%%%%%%%%%%%%%%%%%%%%%%
\subsection{Object Detection and Localization}
%%%%%%%%%%%%%%%%%%%%%%%%%%%%%%%%%%%%%%%%%%%%%%%%%%%

\rmfname~delivers functionality not only to volumetrically explore and map unknown subterranean environments but also to search, detect and localize objects of interest within them. The key elements of this procedure are outlined below.

\subsubsection{Visual Detection}
Object detection is achieved primarily on visual data using a YOLOv3~\cite{yolov3} model trained on a curated dataset of the objects of interest (called ``artifacts''), collected using sensors from aerial and ground platforms using different cameras. Datasets were collected in buildings, abandoned tunnels, mines, natural caves with diverse light conditions and in the presence of obscurants. The object detector was trained using a combined dataset of $40,007$ labels ranging across $8$ classes of objects, namely i) a human survivor, ii) a fire extinguisher, iii) a drill, iv) a backpack, v) a vent, vi) a helmet, vii) a rope, and viii) a cellphone. The trained model is computationally too expensive to be run on the CPU of the Khadas VIM3, however, its integrated $5.0$ TOPS NPU provides dedicated compute for neural network inference. The trained weights file is converted to an NPU compatible format using the provided SDK. The weights of the fully trained neural network were converted from 32-bit floating point numbers to 8-bit integers to run on the NPU, which could then process camera images at $3\textrm{Hz}$. A ROS interface was created to transfer the images from the camera to the NPU, and get detected artifact classes and corresponding bounding boxes as a result. This detection was then used by a multi-view consensus filter to estimate the location of the object.

% Yolov3
% Dataset
% NPU
% Transcoding ?
% Image source?

\begin{figure}[h!]
\centering
    \includegraphics[width=0.99\columnwidth]{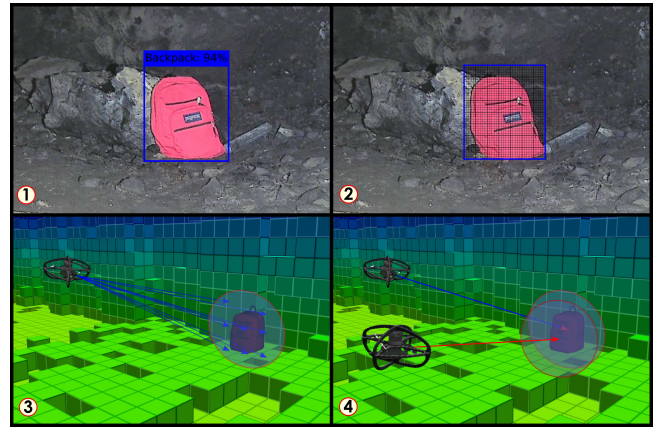}
\caption{Overview of the image detection and localization framework. The artifact is detected onboard (1), the bounding box is divided into pixels (2) and rays are cast into the map (3) to estimate the location of the object. Multiple detections are used over time to robustify the object detection, classification and localization (4).}
\label{fig:objdetection}
\vspace{-2ex} 
\end{figure}

\subsubsection{Localization}
Bounding boxes detected around the artifact are divided into a grid of pixels as shown in Figure \ref{fig:objdetection}. For each of these pixels, rays are cast into the robot's volumetric map $\hat{\boldsymbol{\mu}}_k$ using the camera intrinsic model and the camera-to-LiDAR extrinsic calibration. This results in a set of points that include the object and an area around it. The median point is selected as the estimated position $a_j$ of the object, and a sphere with radius $R_a$ is spawn around it. As the robot moves through the environment, subsequent detections are used to update $a_j$ as an average of all detections projected into the sphere. For each class of object, separate binary Bayesian filters are used to estimate the probability of an object of a certain class being present in the sphere as detailed in~\cite{CERBERUS_SUBT_PHASE_I_II}. Once the probability of a class exceeds a predefined threshold (specific to each class), the process is frozen for the corresponding sphere and the object is reported to the ground station.

\begin{figure*}[h!]
\centering
    \includegraphics[width=0.99\textwidth ]{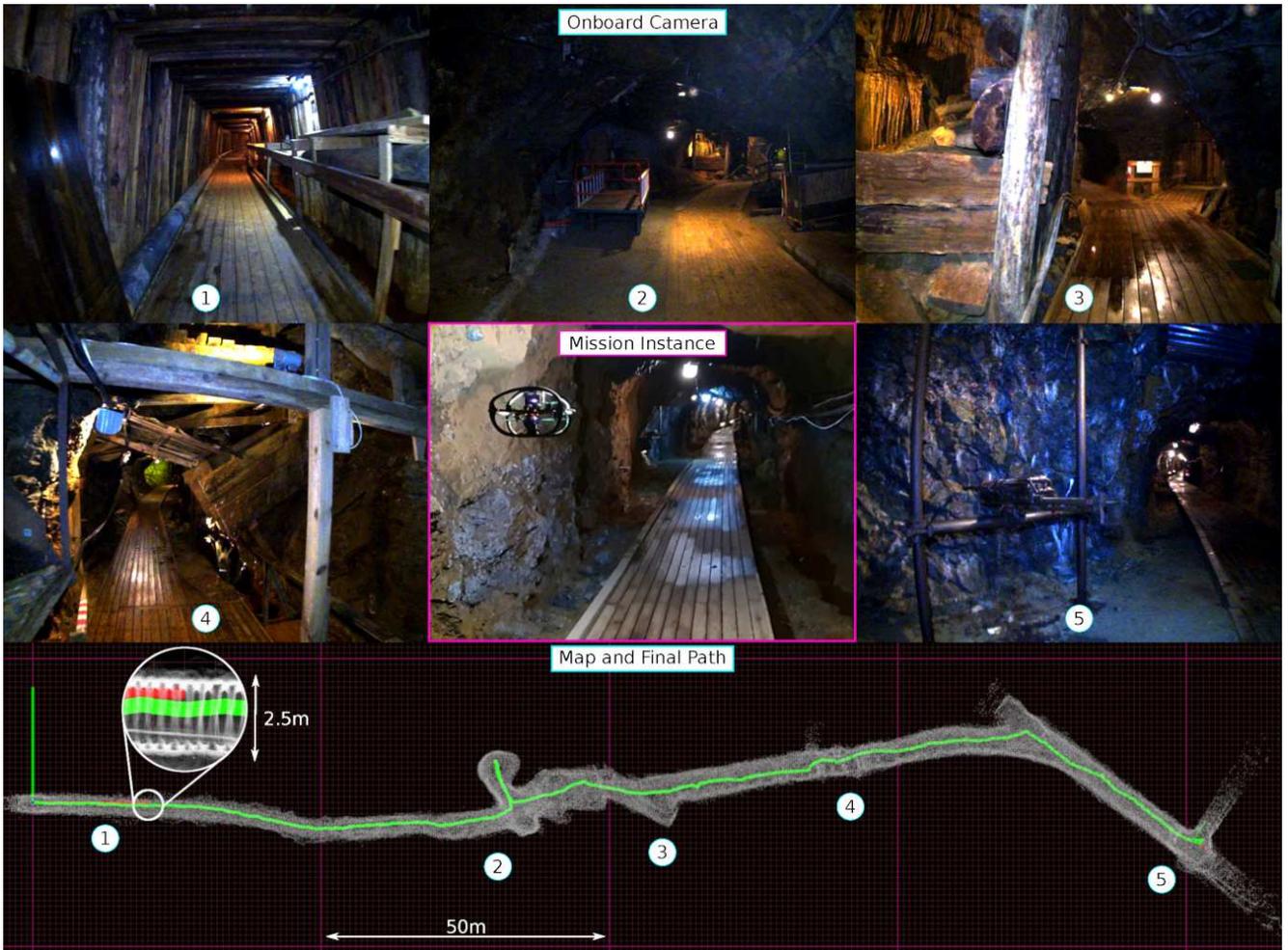} % map_lokken_v3 % 2\columnwidth
    \vspace{-2ex}
\caption{Instances of an autonomous exploration mission inside the L\o kken Mine. (1)-(5) instances from the onboard camera. Central image: \rmfname~ in flight. Bottom row: generated map, final executed path and corresponding location of the upper images. Starting from the main entrance, the robot is able to autonomously takeoff and smoothly plan collision-free paths inside the narrow corridor $(1)$ and reach the first intersection $(2)$. After a quick inspection of a room on the side of the main tunnel, \rmfname~ proceeds its course avoiding other crowded and wet sections $(3)$, $(4)$. Finally, the robot is commanded to safely land in a dry area instead of risking to land inside the drainage channel that is present in the next section of the tunnel $(5)$.}
\label{fig:lokken}
\vspace{-4ex}
\end{figure*}

\subsubsection{Detection of Bluetooth Devices}

A further capability of \rmfname~is to detect devices that have their bluetooth modules on scan mode. Either for devices of known names or in discoverable mode, \rmfname~allows to detect devices such as a cell phone or a computer, while for the location of the device the average robot's location during periods of detecting a certain device is considered. 

\subsubsection{Reporting}
The ground station receives a report containing the class and location of a detected artifact, along with a summarized report of the class probabilities of the artifact. A downsampled image of the detected object is also sent to the ground station. % to verify the correctness of the report by a human supervisor (for visual detections). %The location of the object, with its class is reported to the \textcolor{red}{DARPA server (need non-specific name, or needs prior explanation previously)}, which then evaluates the report and returns information on whether it was correct.

%%%%%%%%%%%%%%%%%%%%%%%%%%%%%%%%%%%%%%%%%%%%%%%%%%%
\subsection{Communications and Networking}
%%%%%%%%%%%%%%%%%%%%%%%%%%%%%%%%%%%%%%%%%%%%%%%%%%%
During mission deployments, \rmfname~can communicate to send data and receive commands through $5\textrm{GHz}$ WiFi. Naturally, in the framework of subterranean exploration communications networking is very hard and the connection to the ground station is expected to be mostly not available. This is also one of the reasons for the emphasis of this work on resilient autonomy. However, when \rmfname~operates in combination with other systems - as for example in the framework of deployments of Team CERBERUS in the DARPA Subterranean Challenge - it is possible to connect to a mesh of WiFi nodes if those are installed, deployed or ferried by other robots as detailed in~\cite{CERBERUS_SUBT_PHASE_I_II}. When such a connection is available, the system shares mapping data, odometry status, battery level, object detection and localization reports, as well as other information. 

%with a base station considered to be outside of the subterranean environment by connecting to the Access Point radio provided by the nodes of a mesh network established using Rajant DX2 ferried by the ground robots of Team CERBERUS, or the Rajant ES1 module on the ground station. All modules operate at $5.8\textrm{GHz}$, with each of the team's legged robot integrating a DX2 module, and our team's roving communications extender also combining one such system with a high gain antenna. This network design is outlined in~\cite{CERBERUS_SUBT_PHASE_I_II} albeit the earlier use of a different communications module. When such a connection is available, the system shares sparsified mapping data, odometry status, battery level, object detection and localization reports, as well as other information. 

%%%%%%%%%%%%%%%%%%%%%%%%%%%%%%%%%%%%%%%%%%%%%%%%%%%%%%%%%%%%%%%%%%%%%%%%%%%%%%%%
\section{FIELD EXPERIMENTS}\label{sec:fie_exp}

This section focuses on selected results from the deployment of \rmfname~in different and challenging environments.

\begin{figure*}[h!]
\centering
    \includegraphics[width=0.99\textwidth]{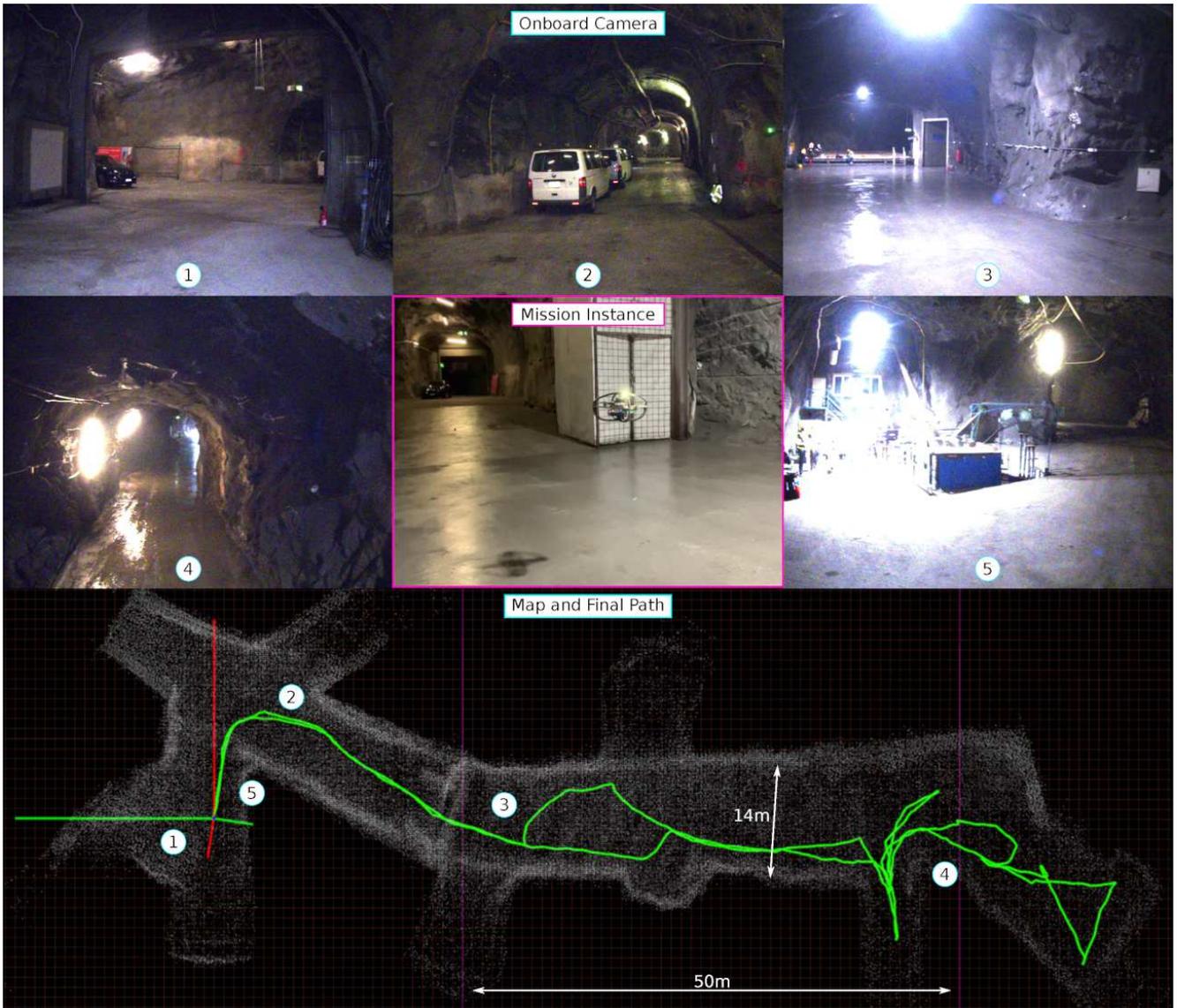} % EPS/map_hagerback_v4.eps % 2\columnwidth
    \vspace{-2ex}
\caption{Instances of an autonomous exploration mission inside the Hagerbach Mine. (1)-(5) instances from the onboard camera. Central image: \rmfname~ in flight. Bottom row: generated map, final executed path and corresponding location of the upper images. Starting from a section of the environment chosen as control station of operation, the robot is able to autonomously takeoff $(1)$,  turn right into the largest branch $(2)$, enter the main room $(3)$ and extensively explore a complete dark section of the mine $(4)$ until the homing path is triggered. Finally \rmfname~safely lands at the takeoff position $(5)$.}
\label{fig:hagerback}
\vspace{-4ex}
\end{figure*}
% \todo{This map contains alignment errors - must be checked. Also caption is incomplete}

%%%%%%%%%%%%%%%%%%%%%%%%%%%%%%%%%%%%%%%%%%%%%%%%%%%%%
\subsection{Løkken Mine}
%%%%%%%%%%%%%%%%%%%%%%%%%%%%%%%%%%%%%%%%%%%%%%%%%%%%%
The first field test took place in the L\o kken Mine, located in the municipality of Orkland in Tr\o ndelag, Norway. This environment was chosen specifically to test the resilience of \rmfname\ in narrow settings. As can be seen in Figure~\ref{fig:lokken}, the first section is less than $2.5\textrm{m}$ wide. The robot has no prior knowledge of the environment and it is commanded to explore as much as possible. From the mine entrance, the robot is able to autonomously takeoff $(1)$, smoothly plan collision-free paths inside the narrow corridor $(2)$ and reach the first intersection $(3)$. After a quick inspection of a room on the side of the main tunnel, \rmfname\ proceeds its course avoiding other wet and crowded sections $(4)$, $(5)$. Finally, the robot is commanded to safely land in a dry area instead of risking landing inside the drainage channel present in the next section of the tunnel. The total travelled distance is more than $200\textrm{m}$ and the total flight time is $6.6\textrm{min}$. Due to the narrow nature of the environment and to permit better tracking performance, the maximum speed was set to $1.0\textrm{m/s}$.

%%%%%%%%%%%%%%%%%%%%%%%%%%%%%%%%%%%%%%%%%%%%%%%%%%%%%
\subsection{Versuchsstollen Hagerbach}
%%%%%%%%%%%%%%%%%%%%%%%%%%%%%%%%%%%%%%%%%%%%%%%%%%%%%
The second field test was conducted in the Versuchsstollen Hagerbach Test Gallery, located in Flums, Switzerland.
In contrast with the previous field test environment, in this case the overall dimensions are considerably bigger, especially if compared to the size of \rmfname. As can be seen in Figure~\ref{fig:hagerback}, the biggest section reaches $14\textrm{m}$ in width. In this second case, the robot also has no prior knowledge of the environment and it is commanded to explore and come back to the takeoff location before the end of its endurance.
Starting from the section chosen as main control station, the robot is able to autonomously takeoff $(1)$, turn right into the largest branch $(2)$, enter the main room $(3)$ and extensively explore a complete dark section of the mine $(4)$ until the homing path is triggered $(5)$. Finally \rmfname~safely lands at the takeoff position $(6)$. Also in this case the total travelled distance is more than $200\textrm{m}$ and the total flight time is $8.1\textrm{min}$, with maximum speed set to $1.2\textrm{m/s}$.

%%%%%%%%%%%%%%%%%%%%%%%%%%%%%%%%%%%%%%%%%%%%%%%%%%%%%%%%%%%%%%%%%%%%%%%%%%%%%%%%
\section{CONCLUSIONS AND FUTURE WORK}\label{sec:concl}
%%%%%%%%%%%%%%%%%%%%%%%%%%%%%%%%%%%%%%%%%%%%%%%%%%%%%%%%%%%%%%%%%%%%%%%%%%%%%%%%
This paper presented the design and overall autonomy functionalities of the \rmfname. Tailored to subterranean exploration, the system emphasizes collision-tolerance and resilient autonomy with robust localization and mapping and high-performance exploration path planning in confined, obstacle-filled, and topologically-complex underground environments. Simultaneously, \rmfname~offers the capacity to search, detect and localize objects of interest which can be proven particularly useful in the framework of search and rescue missions. A set of results from field experiments are presented and allow to demonstrate the capabilities of the system to explore unknown subterranean settings.

\bibliographystyle{IEEEtran}
\bibliography{./RMF-Owl-ICUAS-2022.bib}

\begin{thebibliography}{10}
\providecommand{\url}[1]{#1}
\csname url@samestyle\endcsname
\providecommand{\newblock}{\relax}
\providecommand{\bibinfo}[2]{#2}
\providecommand{\BIBentrySTDinterwordspacing}{\spaceskip=0pt\relax}
\providecommand{\BIBentryALTinterwordstretchfactor}{4}
\providecommand{\BIBentryALTinterwordspacing}{\spaceskip=\fontdimen2\font plus
\BIBentryALTinterwordstretchfactor\fontdimen3\font minus
  \fontdimen4\font\relax}
\providecommand{\BIBforeignlanguage}[2]{{%
\expandafter\ifx\csname l@#1\endcsname\relax
\typeout{** WARNING: IEEEtran.bst: No hyphenation pattern has been}%
\typeout{** loaded for the language `#1'. Using the pattern for}%
\typeout{** the default language instead.}%
\else
\language=\csname l@#1\endcsname
\fi
#2}}
\providecommand{\BIBdecl}{\relax}
\BIBdecl

\bibitem{GBPLANNER_JFR_2020}
T.~Dang, M.~Tranzatto, S.~Khattak, F.~Mascarich, K.~Alexis, and M.~Hutter,
  ``Graph-based subterranean exploration path planning using aerial and legged
  robots,'' \emph{Journal of Field Robotics}, vol.~37, no.~8, pp. 1363--1388,
  2020.

\bibitem{NIR_ICUAS_2017}
{C. Papachristos, S. Khattak and K. Alexis}, ``Autonomous exploration of
  visually-degraded environments using aerial robots,'' in \emph{International
  Conference on Unmanned Aircraft Systems (ICUAS)}.\hskip 1em plus 0.5em minus
  0.4em\relax IEEE, 2017.

\bibitem{sa2014vertical}
I.~Sa and P.~Corke, ``Vertical infrastructure inspection using a quadcopter and
  shared autonomy control,'' in \emph{Field and service robotics}.\hskip 1em
  plus 0.5em minus 0.4em\relax Springer, 2014, pp. 219--232.

\bibitem{bircher2017incremental}
A.~Bircher, K.~Alexis, U.~Schwesinger, S.~Omari, M.~Burri, and R.~Siegwart,
  ``An incremental sampling-based approach to inspection planning: the rapidly
  exploring random tree of trees,'' \emph{Robotica}, vol.~35, no.~6, pp.
  1327--1340, 2017.

\bibitem{tomic2012toward}
T.~Tomic, K.~Schmid, P.~Lutz, A.~Domel, M.~Kassecker, E.~Mair, I.~L. Grixa,
  F.~Ruess, M.~Suppa, and D.~Burschka, ``Toward a fully autonomous uav:
  Research platform for indoor and outdoor urban search and rescue,''
  \emph{IEEE robotics \& automation magazine}, vol.~19, no.~3, pp. 46--56,
  2012.

\bibitem{darpasubt}
\BIBentryALTinterwordspacing
{Defense Advanced Research Projects Agency}, ``{DARPA Subterranean
  Challenge}.'' [Online]. Available: \url{https://subtchallenge.com/}
\BIBentrySTDinterwordspacing

\bibitem{briod2014collision}
A.~Briod, P.~Kornatowski, J.-C. Zufferey, and D.~Floreano, ``A
  collision-resilient flying robot,'' \emph{Journal of Field Robotics},
  vol.~31, no.~4, pp. 496--509, 2014.

\bibitem{salaan2019development}
C.~J. Salaan, K.~Tadakuma, Y.~Okada, Y.~Sakai, K.~Ohno, and S.~Tadokoro,
  ``Development and experimental validation of aerial vehicle with passive
  rotating shell on each rotor,'' \emph{IEEE Robotics and Automation Letters},
  vol.~4, no.~3, pp. 2568--2575, 2019.

\bibitem{atay2021spherical}
S.~Atay, M.~Bryant, and G.~Buckner, ``The spherical rolling-flying vehicle:
  Dynamic modeling and control system design,'' \emph{Journal of Mechanisms and
  Robotics}, vol.~13, no.~5, p. 050901, 2021.

\bibitem{yamada2017development}
M.~Yamada, M.~Nakao, Y.~Hada, and N.~Sawasaki, ``Development and field test of
  novel two-wheeled uav for bridge inspections,'' in \emph{2017 International
  Conference on Unmanned Aircraft Systems (ICUAS)}.\hskip 1em plus 0.5em minus
  0.4em\relax IEEE, 2017, pp. 1014--1021.

\bibitem{RMF_ICRA2021}
P.~D. Petris, H.~Nguyen, M.~Kulkarni, F.~Mascarich, and K.~Alexis, ``Resilient
  collision-tolerant navigation in confined environments,'' in \emph{2021 IEEE
  International Conference on Robotics and Automation (ICRA)}, 2021, pp.
  2286--2292.

\bibitem{sabet2019rollocopter}
S.~Sabet, A.-A. Agha-Mohammadi, A.~Tagliabue, D.~S. Elliott, and P.~E.
  Nikravesh, ``Rollocopter: An energy-aware hybrid aerial-ground mobility for
  extreme terrains,'' in \emph{2019 IEEE Aerospace Conference}.\hskip 1em plus
  0.5em minus 0.4em\relax IEEE, 2019, pp. 1--8.

\bibitem{moon2018uni}
J.-S. Moon, C.~Kim, Y.~Youm, and J.~Bae, ``Uni-copter: A portable
  single-rotor-powered spherical unmanned aerial vehicle (uav) with an
  easy-to-assemble and flexible structure,'' \emph{Journal of Mechanical
  Science and Technology}, vol.~32, no.~5, pp. 2289--2298, 2018.

\bibitem{liu2021toward}
Z.~Liu and K.~Karydis, ``Toward impact-resilient quadrotor design, collision
  characterization and recovery control to sustain flight after collisions,''
  in \emph{2021 IEEE International Conference on Robotics and Automation
  (ICRA)}.\hskip 1em plus 0.5em minus 0.4em\relax IEEE, 2021, pp. 183--189.

\bibitem{mintchev2017insect}
S.~Mintchev, S.~de~Rivaz, and D.~Floreano, ``Insect-inspired mechanical
  resilience for multicopters,'' \emph{IEEE Robotics and automation letters},
  vol.~2, no.~3, pp. 1248--1255, 2017.

\bibitem{klaptocz2013euler}
A.~Klaptocz, A.~Briod, L.~Daler, J.-C. Zufferey, and D.~Floreano, ``Euler
  spring collision protection for flying robots,'' in \emph{2013 IEEE/RSJ
  International Conference on Intelligent Robots and Systems}.\hskip 1em plus
  0.5em minus 0.4em\relax IEEE, 2013, pp. 1886--1892.

\bibitem{sareh2018rotorigami}
P.~Sareh, P.~Chermprayong, M.~Emmanuelli, H.~Nadeem, and M.~Kovac,
  ``Rotorigami: A rotary origami protective system for robotic rotorcraft,''
  \emph{Science Robotics}, vol.~3, no.~22, 2018.

\bibitem{kornatowski2017origami}
P.~M. Kornatowski, S.~Mintchev, and D.~Floreano, ``An origami-inspired cargo
  drone,'' in \emph{2017 IEEE/RSJ International Conference on Intelligent
  Robots and Systems (IROS)}.\hskip 1em plus 0.5em minus 0.4em\relax IEEE,
  2017, pp. 6855--6862.

\bibitem{sareh2018spinning}
P.~Sareh, P.~Chermprayong, M.~Emmanuelli, H.~Nadeem, and M.~Kovac, ``The
  spinning cyclic ‘miura-oring’for mechanical collision-resilience,''
  \emph{Origami 7}, vol.~3, pp. 981--994, 2018.

\bibitem{zha2020collision}
J.~Zha, X.~Wu, J.~Kroeger, N.~Perez, and M.~W. Mueller, ``A collision-resilient
  aerial vehicle with icosahedron tensegrity structure,'' in \emph{2020
  IEEE/RSJ International Conference on Intelligent Robots and Systems
  (IROS)}.\hskip 1em plus 0.5em minus 0.4em\relax IEEE, 2020, pp. 1407--1412.

\bibitem{de2021being}
R.~de~Azambuja, H.~Fouad, Y.~Bouteiller, C.~Sol, and G.~Beltrame, ``When being
  soft makes you tough: A collision-resilient quadcopter inspired by
  arthtropods' exoskeletons,'' \emph{arXiv preprint arXiv:2103.04423}, 2021.

\bibitem{klaptocz2010indoor}
A.~Klaptocz, G.~Boutinard-Rouelle, A.~Briod, J.-C. Zufferey, and D.~Floreano,
  ``An indoor flying platform with collision robustness and self-recovery,'' in
  \emph{2010 IEEE International Conference on Robotics and Automation}.\hskip
  1em plus 0.5em minus 0.4em\relax IEEE, 2010, pp. 3349--3354.

\bibitem{vourtsis2021robotic}
C.~Vourtsis, W.~Stewart, and D.~Floreano, ``Robotic elytra: Insect-inspired
  protective wings for resilient and multi-modal drones,'' \emph{IEEE Robotics
  and Automation Letters}, 2021.

\bibitem{phan2020mechanisms}
H.~V. Phan and H.~C. Park, ``Mechanisms of collision recovery in flying beetles
  and flapping-wing robots,'' \emph{Science}, vol. 370, no. 6521, 2020.

\bibitem{chen2021collision}
Y.~Chen, S.~Xu, Z.~Ren, and P.~Chirarattananon, ``Collision resilient
  insect-scale soft-actuated aerial robots with high agility,'' \emph{IEEE
  Transactions on Robotics}, 2021.

\bibitem{zha2021exploiting}
J.~Zha and M.~W. Mueller, ``Exploiting collisions for sampling-based
  multicopter motion planning,'' in \emph{2021 IEEE International Conference on
  Robotics and Automation (ICRA)}.\hskip 1em plus 0.5em minus 0.4em\relax IEEE,
  2021, pp. 7943--7949.

\bibitem{mote2016framework}
M.~L. Mote, J.-P. Afman, and E.~Feron, ``A framework for collision-tolerant
  optimal trajectory planning of autonomous vehicles,'' \emph{arXiv preprint
  arXiv:1611.07608}, 2016.

\bibitem{liu2021sensing}
C.~Liu and R.~Tron, ``Sensing via collisions: a smart cage for quadrotors with
  applications to self-localization,'' in \emph{2021 IEEE International
  Conference on Robotics and Automation (ICRA)}.\hskip 1em plus 0.5em minus
  0.4em\relax IEEE, 2021.

\bibitem{lew2019contact}
T.~Lew, T.~Emmei, D.~D. Fan, T.~Bartlett, A.~Santamaria-Navarro, R.~Thakker,
  and A.-a. Agha-mohammadi, ``Contact inertial odometry: Collisions are your
  friends,'' \emph{arXiv preprint arXiv:1909.00079}, 2019.

\bibitem{mulgaonkar2020tiercel}
Y.~Mulgaonkar, W.~Liu, D.~Thakur, K.~Daniilidis, C.~J. Taylor, and V.~Kumar,
  ``The tiercel: A novel autonomous micro aerial vehicle that can map the
  environment by flying into obstacles,'' in \emph{2020 IEEE International
  Conference on Robotics and Automation (ICRA)}.\hskip 1em plus 0.5em minus
  0.4em\relax IEEE, 2020, pp. 7448--7454.

\bibitem{mulgaonkar2017robust}
Y.~Mulgaonkar, A.~Makineni, L.~Guerrero-Bonilla, and V.~Kumar, ``Robust aerial
  robot swarms without collision avoidance,'' \emph{IEEE Robotics and
  Automation Letters}, vol.~3, no.~1, pp. 596--603, 2017.

\bibitem{RADMF_ICRA_2021}
F.~Mascarich, P.~De~Petris, H.~Nguyen, and K.~Alexis, ``Autonomous distributed
  3d radiation field estimation for nuclear environment characterization,'' in
  \emph{2021 IEEE International Conference on Robotics and Automation
  (ICRA)}.\hskip 1em plus 0.5em minus 0.4em\relax IEEE, 2021.

\bibitem{CompSLAM}
S.~Khattak, H.~Nguyen, F.~Mascarich, T.~Dang, and K.~Alexis, ``Complementary
  multi–modal sensor fusion for resilient robot pose estimation in
  subterranean environments,'' in \emph{2020 International Conference on
  Unmanned Aircraft Systems (ICUAS)}, 2020, pp. 1024--1029.

\bibitem{loam}
J.~Zhang and S.~Singh, ``Loam: Lidar odometry and mapping in real-time.'' in
  \emph{Robotics: Science and Systems}, vol.~2, no.~9, 2014.

\bibitem{lynen13robust}
S.~Lynen, M.~Achtelik, S.~Weiss, M.~Chli, and R.~Siegwart, ``A robust and
  modular multi-sensor fusion approach applied to mav navigation,'' in
  \emph{Proc. of the IEEE/RSJ Conference on Intelligent Robots and Systems
  (IROS)}, 2013.

\bibitem{voxblox}
H.~Oleynikova, Z.~Taylor, M.~Fehr, R.~Siegwart, and J.~Nieto, ``Voxblox:
  Incremental 3d euclidean signed distance fields for on-board mav planning,''
  in \emph{IEEE/RSJ International Conference on Intelligent Robots and Systems
  (IROS)}, 2017.

\bibitem{GBPlanner2_ArxivVersion}
M.~Kulkarni, M.~Dharmadhikari, M.~Tranzatto, S.~Zimmermann, V.~Reijgwart,
  P.~De~Petris, H.~Nguyen, N.~Khedekar, C.~Papachristos, L.~Ott, R.~Siegwart,
  M.~Hutter, and K.~Alexis, ``Autonomous teamed exploration of subterranean
  environments using legged and aerial robots,'' \emph{arXiv preprint
  arXiv:2111.06482}, 2021.

\bibitem{michael2011cooperative}
N.~Michael, J.~Fink, and V.~Kumar, ``Cooperative manipulation and
  transportation with aerial robots,'' \emph{Autonomous Robots}, vol.~30, 2011.

\bibitem{yolov3}
J.~Redmon and A.~Farhadi, ``Yolov3: An incremental improvement,'' \emph{arXiv},
  2018.

\bibitem{CERBERUS_SUBT_PHASE_I_II}
M.~Tranzatto, F.~Mascarich, L.~Bernreiter, C.~Godinho, M.~Camurri, S.~M.~K.
  Khattak, T.~Dang, V.~Reijgwart, J.~Loeje, D.~Wisth, , S.~Zimmermann,
  H.~Nguyen, M.~Fehr, L.~Solanka, R.~Buchanan, M.~Bjelonic, N.~Khedekar,
  M.~Valceschini, F.~Jenelten, M.~Dharmadhikari, T.~Homberger, P.~De~Petris,
  L.~Wellhausen, M.~Kulkarni, T.~Miki, S.~Hirsch, M.~Montenegro,
  C.~Papachristos, F.~Tresoldi, J.~Carius, G.~Valsecchi, J.~Lee, K.~Meyer,
  X.~Wu, J.~Nieto, A.~Smith, M.~Hutter, R.~Siegwart, M.~Mueller, M.~Fallon, and
  K.~Alexis, ``Cerberus: Autonomous legged and aerial robotic exploration in
  the tunnel and urban circuits of the darpa subterranean challenge,''
  \emph{Journal of Field Robotics}, 2021.

\end{thebibliography}

\end{document}